\documentclass{llncs}

\usepackage[T1]{fontenc}
\usepackage[english]{babel}
\usepackage{graphicx}
\usepackage{amsmath}
\usepackage{multirow}
\usepackage{booktabs}

\usepackage{xcolor}
\usepackage{wrapfig}

\usepackage{natbib}
\begin{document}

\title{Physics-Guided Regime Unmixing}

\titlerunning{Physics-Guided Regime Unmixing}

\author{
    Paula Pacheco\inst{1,2,3}\orcidID{0009-0005-0005-1515},
    Pablo Granitto\inst{4}\orcidID{0000-0002-2473-8769},
    Juan B. Cabral\inst{1,2,3}\orcidID{0000-0002-7351-0680}\\
    \email{paulapacheco@mi.unc.edu.ar}
}

\authorrunning{Pacheco et al.}

\institute{
    GVT-CONAE, Centro Espacial Teófilo Tabanera, Argentina
\and
    CONICET, Argentina
\and
    FAMAF-UNC, Córdoba, Argentina
\and
    CIFASIS, CONICET--UNR, Rosario, Argentina
}

\maketitle

\begin{abstract}
The Linear Mixing Model (LMM) dominates spectral unmixing for its simplicity, but fails under multiple scattering; existing nonlinear models compensate by applying a fixed regime uniformly across entire scenes. We propose \emph{Physics-Guided Regime Unmixing} (PGRU), which estimates a pixel-wise scalar $\xi_i \in [0,1]$ from observable physical features to activate nonlinear mixing only where justified. Residuals from the Generalized Bilinear Model (GBM), the Post-Nonlinear Mixing Model (PPNM), and Hapke are combined via learned attention, yielding interpretable regime maps. Experiments on Samson, Jasper Ridge, and Urban show consistent improvements over baselines, with physical coherence $\rho > 0.90$.

\end{abstract}

\begin{center}
\vspace{.5cm}
{\Large \bfseries\boldmath Régimen de Desmezclado Guiado por Física}
\vspace{0.5cm}
\end{center}

\begin{abstract}
El Modelo de Mezcla Lineal (LMM) predomina en el desmezclado espectral por su simplicidad, pero falla ante la dispersión múltiple; los modelos no lineales existentes compensan esto aplicando un régimen fijo de manera uniforme en toda la escena. Proponemos el \emph{Physics-Guided Regime Unmixing} (PGRU), que estima un escalar por píxel $\xi_i \in [0,1]$ a partir de características físicas observables para activar la mezcla no lineal únicamente donde está justificado. Los residuos del Modelo Bilineal Generalizado (GBM), el Modelo de Mezcla Post-No Lineal (PPNM) y Hapke se combinan mediante atención aprendida, 
generando mapas de régimen interpretables. Los experimentos en Samson, Jasper Ridge y Urban muestran mejoras consistentes frente a las líneas base, con coherencia física $\rho > 0.90$.

\end{abstract}

\vspace{0.5cm}
\setcounter{page}{1}

\newpage
\section{Introduction}

Hyperspectral sensors capture surface reflectance across hundreds of contiguous spectral bands, enabling fine-grained material characterization for applications including agriculture, geology, and environmental monitoring~\citep{bhargava2024hyperspectral}. Because spatial resolution is typically coarser than the scale of surface heterogeneity, each pixel contains the mixed spectral response of multiple materials. Spectral unmixing recovers the pure signatures (endmembers) and their fractional contributions (abundances) from this mixture.

The LMM~\citep{keshava2002spectral} is the standard approach due to its simplicity and physical interpretability, but assumes photons interact with a single material before reaching the sensor. This assumption breaks down under multiple scattering, motivating nonlinear extensions such as the Generalized Bilinear Model (GBM)~\citep{halimi2011nonlinear}, the Post-Nonlinear Mixing Model (PPNM)~\citep{altmann2012supervised}, and the Hapke radiative transfer model~\citep{hapke1981bidirectional}. These models improve reconstruction accuracy in strongly nonlinear scenes, but apply nonlinearity uniformly across all pixels, without distinguishing which pixels actually require it or why.

Deep learning approaches, primarily autoencoders and convolutional networks~\citep{palsson2018hyperspectral,palsson2022blind}, achieve strong reconstruction performance, but operate as black boxes and offer no physical explanation for regime selection.

We address this gap with PGRU, which introduces a pixel-wise regime parameter $\xi_i \in [0,1]$ guided by observable physical features. Unlike previous approaches, PGRU does not require labels indicating whether pixels are linear or nonlinear: regime assignment emerges from the tension between reconstruction improvement and physical feature consistency.

\section{Physics-Guided Regime Unmixing}

Each pixel $i$ is reconstructed as $\hat{y}_i = S_{\mathrm{lin},i} + \xi_i\,\delta_{\mathrm{nl},i}$, where $S_{\mathrm{lin},i} = \sum_m a_{im} e_m$ is the standard LMM component (non-negative abundances summing to one) and $\delta_{\mathrm{nl},i}$ is a nonlinear residual weighted by $\xi_i \in [0,1]$. When $\xi_i \approx 0$ the model reduces to LMM; when $\xi_i \approx 1$ the nonlinear contribution is fully active.

The residual $\delta_{\mathrm{nl},i} = \sum_k \alpha_{ik}\,\delta_{ik}$ combines three physical models via learned attention weights $\alpha_{ik}$ obtained through a temperature-controlled softmax: GBM captures pairwise scattering between endmembers, PPNM introduces quadratic distortion over the linear mixture, and Hapke describes multiple scattering from radiative transfer theory. Entropy regularization on $\alpha_{ik}$ encourages concentration toward the dominant physical mechanism at each pixel.

The central contribution is that $\xi_i$ is guided by observable scene properties rather than driven solely by reconstruction error.

\begin{equation}
  \xi_i = \sigma\!\left(\sum_k w_k F_{ik} + b\right)
\end{equation}

where $F_{ik}$ are physical and geometric features of pixel $i$ (spectral curvature, NDVI, NDVI gradient, EMP, DMP, LBP), $w_k$ are learned weights, and $\sigma$ is the sigmoid. This is equivalent to a logistic regression over the feature space: the sign and magnitude of $w_k$ directly reveal which scene properties predict nonlinear behavior and in which direction, making the regime decision for any pixel a readable sum of feature contributions.

To train the model, a per-pixel reconstruction gain is defined as
$\Delta_{\mathrm{res},i} = \|y_i - S_{\mathrm{lin},i}\| -
\|y_i - \hat{y}_i\|$, and the full objective is:

\begin{equation}
  \mathcal{L} = \sum_i \Bigl[
    -\tanh(\Delta_{\mathrm{res},i})
    + \lambda_{\mathrm{feat}}(\xi_i - \xi_{\mathrm{prior},i})^2
    + \lambda_{\mathrm{sp}}\,\xi_i \nabla^2 \xi_i
  \Bigr] + \lambda_w \|w\|^2
\end{equation}

The first term rewards nonlinear activation where it reduces reconstruction error, saturated by $\tanh$ for large gains. The second anchors $\xi_i$ to the feature-based prior, with $\lambda_{\mathrm{feat}}$ annealed during training. The third promotes spatial coherence of the regime map via a Laplacian
regularizer. No pixel labels are used at any stage: the regime assignment emerges entirely from the interplay between reconstruction evidence and physical feature consistency.

\section{Experiments}


We evaluate PGRU on three widely used benchmark datasets~\citep{zhu2017hyperspectral}: \textbf{Samson} ($95\!\times\!95$ pixels, 156 bands, 3 endmembers: water, vegetation, soil), \textbf{Jasper Ridge} ($100\!\times\!100$, 198 bands, 4 endmembers: water, trees, soil, road), and \textbf{Urban} ($307\!\times\!307$, 162 bands, 5 endmembers: asphalt, grass, trees, roof, soil). Reference endmembers provided with each dataset are used in all methods. Baselines are LMM (linear), GBM, and PPNM; all methods share the same abundance estimation procedure. Reconstruction is evaluated with Spectral Angle Distance (SAD), RMSE, and relative RMSE (rRMSE).


Table~\ref{tab:results} reports reconstruction metrics across all
datasets and methods.
Preliminary experiments indicate that PGRU achieves lower reconstruction errors on all three datasets.
The improvement is especially pronounced in Samson (rRMSE drops from
0.495 to 0.052) and Jasper Ridge (rRMSE from 0.347 to 0.068), with
substantial gains also observed in the more complex Urban scene.
Notably, the nonlinear baselines (GBM, PPNM) do not consistently
outperform LMM, confirming that applying nonlinearity uniformly can
be counterproductive.
PGRU avoids this by activating nonlinearity only where justified.

\begin{table}[t]
\centering
\caption{Reconstruction metrics on three benchmark datasets.
Best results in \textbf{bold}.}
\label{tab:results}
\begin{tabular}{llcccc}
\toprule
\textbf{Dataset} & \textbf{Metric} &
\textbf{LMM} & \textbf{GBM} & \textbf{PPNM} & \textbf{PGRU} \\
\midrule
\multirow{3}{*}{Samson}
 & SAD   & 0.158 & 0.141 & 0.143 & \textbf{0.052} \\
 & RMSE  & 0.121 & 0.141 & 0.132 & \textbf{0.013} \\
 & rRMSE & 0.495 & 0.578 & 0.540 & \textbf{0.052} \\
\midrule
\multirow{3}{*}{Jasper Ridge}
 & SAD   & 0.288 & 0.290 & 0.287 & \textbf{0.151} \\
 & RMSE  & 0.055 & 0.075 & 0.070 & \textbf{0.011} \\
 & rRMSE & 0.347 & 0.476 & 0.444 & \textbf{0.068} \\
\midrule
\multirow{3}{*}{Urban}
 & SAD   & 0.121 & 0.120 & 0.115 & \textbf{0.070} \\
 & RMSE  & 0.067 & 0.087 & 0.081 & \textbf{0.013} \\
 & rRMSE & 0.331 & 0.430 & 0.400 & \textbf{0.064} \\
\bottomrule
\end{tabular}
\end{table}


Beyond reconstruction accuracy, we evaluate whether the regime
activation maps are physically grounded. Physical coherence $\rho = \mathrm{corr}(\xi, \Delta_{\mathrm{res}})$ measures the correlation between the activation map and the per-pixel
reconstruction gain. Values exceed 0.90 across all datasets ($\rho = 0.93$, $0.91$, $0.92$ for Samson, Jasper Ridge, and Urban respectively), confirming that PGRU activates nonlinear mixing in regions where reconstruction improvement is observed.

The feature weight vector $w$ reveals which physical properties predict
nonlinear behavior in each scene. In Jasper Ridge, NDVI dominates: nonlinear activation is concentrated in dense vegetation, where multiple scattering is expected, and suppressed over the water body. In Urban, spectral curvature drives regime selection, reflecting structural complexity at material boundaries rather than vegetation-related interactions. This scene-adaptive behavior emerges without supervision.

\begin{figure}[t]
  \centering
  \includegraphics[width=\textwidth]{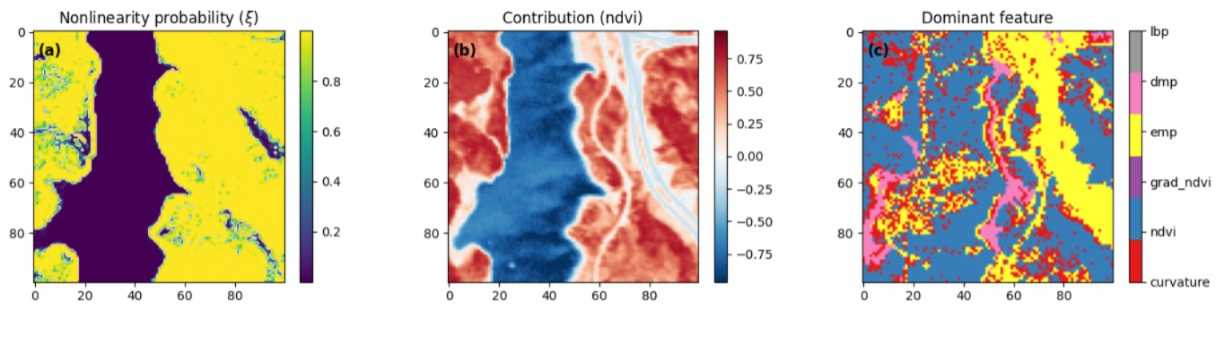}
  \caption{Jasper Ridge: (a) nonlinearity map $\xi$, (b) NDVI contribution,
  (c) dominant feature per pixel.
  Nonlinear activation concentrates in vegetated areas, consistent
  with expected multiple-scattering behavior.}
  \label{fig:maps}
\end{figure}

Beyond global feature importance, PGRU provides pixel-level explainability through spatial maps that decompose regime selection into feature contributions.
Fig.~\ref{fig:maps} illustrates this behavior for Jasper Ridge.
The spatial organization of these explanations suggests that the model does not arbitrarily assign regimes, but anchors them to measurable scene properties, providing direct answers to the question \emph{why is this pixel nonlinear?}

\section{Conclusion}

We presented PGRU, which addresses a question often left open by prior unmixing methods: \emph{why is this pixel nonlinear?} By grounding the regime scalar $\xi_i$ in observable physical features, the model produces interpretable regime maps without any labeled data. Initial results on three benchmarks suggest that selective, explainable nonlinearity outperforms both uniform linear and uniform nonlinear approaches. These results are preliminary and require broader validation against recent deep-learning baselines. A current limitation is that the contribution of each component has not been validated in isolation: the relative importance of the physical model ensemble, the feature-guided activation of $\xi_i$, and the
spatial regularization remain to be disentangled through a systematic ablation study. Future work will address this limitation alongside several extensions. In particular, we plan to extend PGRU to blind unmixing settings and to scenarios with ground-truth nonlinear interaction labels. We also aim to incorporate spatial context into regime estimation, explore alternative feature sets, and evaluate the method on more complex nonlinear mixing scenarios, including comparisons with recent deep learning-based unmixing approaches. 
PGRU suggests that selective, interpretable nonlinearity can be a practical alternative to globally linear or globally nonlinear unmixing models.

\begin{credits}
\subsubsection{\ackname}

This work was partially supported by CONICET (Argentina). P.P. was supported by a fellowship from CONICET and CONAE.

\noindent Language editing assistance was provided by AI tools. All scientific content remains the responsibility of the authors.
\end{credits}

\bibliographystyle{plainnat}
\bibliography{thebiblio}

\end{document}